# Deep Learning Image Recognition for Non-images

Boris Kovalerchuk[1], Divya Chandrika Kalla[1], Bedant Agarwal[2]

[1]Dept. of Computer Science, Central Washington University, USA
borisk@cwu.edu, DivyaChandrika.Kalla@cwu.edu
[2] Indian Institute of Technology Kharagpur, India
BedantAgarwal9@gmail.com

**Abstract**. Powerful deep learning algorithms open an opportunity for solving non-image Machine Learning (ML) problems by transforming these problems into the image recognition problems. The CPC-R algorithm presented in this chapter converts non-image data into images by visualizing non-image data. Then deep learning CNN algorithms solve the learning problems on these images. The design of the CPC-R algorithm allows preserving all high-dimensional information in 2-D images. The use of pair values mapping instead of single value mapping used in the alternative approaches allows encoding each n-D point with 2 times fewer visual elements. The attributes of an n-D point are divided into pairs of its values and each pair is visualized as 2-D points in the same 2-D Cartesian coordinates. Next, grey scale or color intensity values are assigned to each pair to encode the order of pairs. This is resulted in the heatmap image. The computational experiments with CPC-R are conducted for different CNN architectures, and methods to optimize the CPC-R images showing that the combined CPC-R and deep learning CNN algorithms are able to solve non-image ML problems reaching high accuracy on the benchmark datasets. This chapter expands our prior work by adding more experiments to test accuracy of classification, exploring saliency and informativeness of discovered features to test their interpretability, and generalizing the approach.

## 1. Introduction

The success in solving image recognition problems by Deep Learning (DL) algorithms is very evident. Moreover, DL architectures designed for some types of images have been efficient for other types of images. This chapter expands such knowledge transfer opportunity by converting non-image data to images by using visualization of non-image data. It can solve a wide variety of Machine Learning problems [4,10] by converting a non-image classification task into the image recognition task and solving it by efficient DL algorithms. This chapter expands [26] as follows: (1) adding more experiments with more benchmark datasets to test accuracy of classification, (2) exploring saliency and informativeness of discovered features to test their interpretability, and (3) generalizing the approach.

This chapter starts from the concepts of single value and pair value mappings. Sections 2-4 present result of experiments with Wisconsin breast cancer data (WBC), Swiss roll data, Ionosphere, Glass, and Car data. Section 5 presents results with Saliency maps and Section 6 presents results with Informative cells feature importance model. Section 7 explores advantages of Spiral filling of colliding pairs. Section 8 presents results for frequencies of informative cells. Section 9 presents comparisons with other studies. Generalization of CPC-R methodology, future work and conclusion are presented in Sections 10 and 11.

## 1.1. Single value mapping

In a *single value mapping* [26], each *single* value $a_i$ of the n-D point $\mathbf{a}=(a_1,a_2,\ldots,a_n)$ is mapped by function $F$ to a cell/pixel $p_i$ with coordinates $(p_{i1},p_{i2})$, $F(a_i)=(p_{i1},p_{i2})$. The coordinate value $a_i$ is assigned as the intensity of pixel $p_i$. Before the value $a_i$ can be normalized to be an intensity value, $I(p_i)=a_i$. Next, we define the *mapping density* relative to the image size. For $n=3$, consider the image of size 2×2 pixels with mapping $a_1$ to pixel (1,1), $a_2$ to pixel (1,2) and $a_3$ to pixel (2,1) with pixel (2,2) left empty (see Fig 1ab). Such sequential mapping is a dense mapping, e.g., it needs only 10×10 pixels for 100-D point **a**. The use of a larger 3×3 image with $F(a_1)=(1,1)$, $F(a_2)=(2,2)$, and $F(a_3)=(3,3)$ leads to density 3/9, where these 3 pixels contain relevant information. See Fig. 1cd.

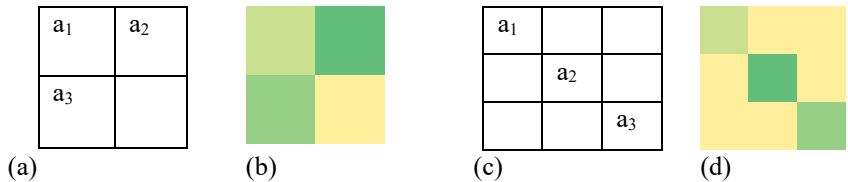

(a)        (b)        (c)        (d)
Fig. 1. Sequential (a,b) and diagonal (c,d) mappings.

These examples illustrate only a few options to map all attributes of **a** to pixels in the $k\times k$ pixels image, while multiple other mappings are possible. It leads to the *optimization task* of optimal mapping attributes of **a** to pixels that will maximize the *classification accuracy* by a given machine learning algorithm. The formal mathematical exploration of this task still is a future task. The practical approach taken in [10] is discovering relations between attributes $a_i$ and putting related attributes to adjacent pixels avoiding exhaustive search of the optimal mapping. Its novelty is that PCA, K-PCA or t-SNE mapping is used not as a visualization itself, but only to identify the *location* of the pixels on the 2-D plot. While, in general, it is productive approach, a specific relation discovered by such *unsupervised* methods can be irrelevant to the supervised classification task. Next, the distances that PCA, and t-SNE discover to locate pixels nearby can corrupt n-D distances [13].

Recently, a single value mapping has been successfully applied to timeseries [39], where different univariate time series for the same sample are displayed on the same plot and run by a single CNN. The points can be connected to form a graph. Alternatively, the univariate time series are displayed in separate plots and run by several "Siamese" CNNs with the same convolutional and pooling layers. Then before flattening, the outputs of these CNNs are joined and run by the same classifier section to produce a single output.

*1.2. Graph mappings*

The algorithm GLC-L in [4] represents an n-D point **a** as a 2-D *graph*, which is a sequence of connected straight lines under different angles where line $L_i$ has the length of $a_i$ of n-D point **a**. The disadvantage of this approach is the *size* of its raster image because drawing graph edges requires many pixels. Methods in [4,10,39] use images that they produce as inputs to CNN algorithms for *image recognition* and all reported success on different datasets. Their disadvantage is a single value mapping requiring a cell (pixel) for each $a_i$ of n-D point **a**. In general, single value mappings can produce a *sparce mapping* where only a small fraction of pixels represent each n-D point **a**. It is illustrated in Figs. 2 and 3, which show other single-value mapping options. Fig. 2a shows an option from [17], which encodes each value $a_i$ as bars of (max-$a_i$) going from the top that we call a *reversed bar*. Fig. 2b shows an equivalent single-value mapping using *parallel coordinates* for the same max-$a_i$ values. In general, it is possible to map each $a_i$ itself without making max-$a_i$. Fig. 3 presents 5-D point **a** =(2, 4,5,1,6) in lossless GLC-L and Collocated Paired Coordinates (CPC) [35].

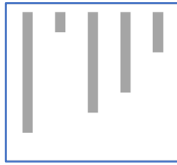 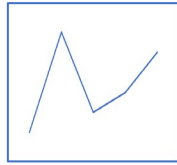 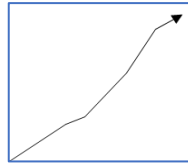 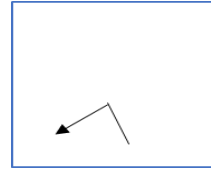

(a)            (b)            (a)            (b)

Fig. 2. 5-D point **x**=($x_1$, $x_2$, $x_3$, $x_4$,$x_5$)= (6,1,5,4,2) in reversed bar graph (a), i.e., showing bars of (max-$a_i$) going from the top, and (b) in parallel coordinates.

Fig. 3. 5-D point **a** =(2, 4,5,1,6) in GLC-L (a) and Collocated Paired Coordinates (b).

Fig. 4 shows *partial pair values mappings* [17], where Fig. 4a shows a heatmap of differences $x_i$-$x_j$ and Fig 4b uses this heatmap as a background for the reversed bar chart shown in Fig.2a.

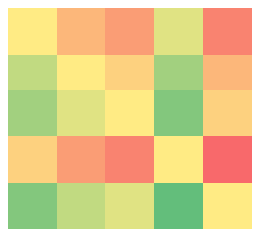 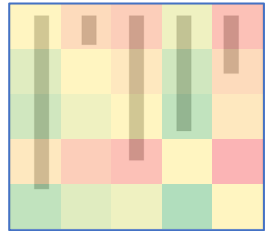

(a)            (b)

Fig. 4. (a) Heatmap of differences $x_i$-$x_j$ used as a background for (b) that is a reversed bar chart.

The proposed **algorithm CPC-R** uses a *pair values mapping with preserving* all n-D information. It has advantages over single value mappings such as proposed in [4]: (1) *two times less* cells (pixels), (2) smaller images, and (3) less occlusion because CPC-R does not need the graph edges.

## 1.3. Pair values mapping

A *pair values mapping H* maps a pair of values $(a_i, a_j)$ to a single pixel $p$, $H(a_i, a_j) = (p_{1ij}, p_{2ij})$. Thus, the values $a_i$ and $a_j$ are encoded by the *location* of the pixel. In a simple version of $H$ mapping, the coordinates of the pixels are values $a_i$ and $a_j$, $(p_{1ij}, p_{2ij}) = (a_i, a_j)$ and $(p_{1i,i+1}, p_{2i,i+1}) = (a_i, a_{i+1})$. It puts similar pairs next to each other. This requires converting $a_i$ and $a_j$ to integers by *rounding* them. When $n$ is odd we form the last pair $(a_n, a_n)$ by repeating $a_n$. The *grey scale intensities* of the pixels are used to specify the order of the pairs, where the first pair is black; the second pair is dark grey and so on. The last pair is very light grey. A color sequence is also used as an alternative producing a *heatmap*. The values of intensities and colors can be optimized at the learning stage. We denote this mapping algorithm as **CPC-R algorithm** and images that it produces as **CPC-R images**, because it is inspired by visualization of n-D points by *Collocated Paired Coordinates* (CPC) in vector graphics [35]. The letter R indicates that CPC-R are raster non vector images.

The benefit of the CPC-R pair values mapping is in representing an n-D point *without loss of information* by using only $\lceil n/2 \rceil$ pixels, which is two times less pixels than in a single pixel mapping. It makes images *simpler*. The CPC-R algorithm allows to optimize the location of the pixels by using alternative pairing $(a_i, a_j)$ not only $(a_i, a_{i+1})$ to increase classification accuracy. A simplest version of CPC-R algorithm tests a fixed number of alternative pairings randomly generated with a given classification algorithm.

## 1.4. Pair values mapping algorithm CPC-R

A large part of this chapter presents the experiments to classify the real world and simulated data with the CPC-R and CNN algorithms. We tested the efficiency of CPC-R+CNN algorithm successfully by evaluating the accuracy of the classification on images produced from benchmark datasets that have been discretized to map to the CPC-R images.

The steps of the **base CPC-R algorithm version 1.0** are as follow:

S1. *Split attributes* of an n-D point $\mathbf{x} = (x_1, x_2, \ldots, x_n)$ to consecutive pairs $(x_1, x_2)$, $(x_3, x_4)$, …, $(x_{n-1}, x_n)$. If $n$ is odd repeat the last attribute of $\mathbf{x}$ to make $n$ even.

S2. *Set up a cell size* to locate pairs in the image. Each cell can consist of a single pixel or dozens of pixels.

S3. *Locate pairs* $(x_i, x_{i+1})$ in the image at the cell coordinate pair $(x_i, x_{i+1})$ in the image.

S4. *Assign the grey scale intensity* to cells from black for $(x_1, x_2)$ and very light grey for $(x_{n-1}, x_n)$. Alternatively, *assign heatmap colors*.

S5. *Optional: Combine* produced images with context information in the form of average images of classes.

S6. *Call/Run ML/DL* algorithm to discover a predictive model.

S7. *Optional: Optimize* intensities in S4 and pairing coordinates beyond the sequential pairs ($x_i$, $x_{i+1}$) used in S1 by the methods ranged from random generating a fixed number of alternatives to genetic algorithms with testing ML prediction accuracy for these alternatives.

Full restoration of the order of the pairs and their values from a CPC-R image is possible due to the order of intensities if all pairs differ. The equal pairs will *collide* in the image location. The base algorithm keeps only the intensity of the *last pair* from a set of colliding pairs. Other versions preserve more information about the colliding pairs.

According to this algorithm a small image with 10x10 pixels can represent a 10-D point, when each attribute has 10 different values, and each cell uses only a single pixel in step 2. It will locate five grey scale pixels in this image.

The generation of raster images in CPC-R for the Wisconsin Breast Cancer (WBC) data [41] is illustrated in Fig. 5. We consider a 10-D point **x**=(8,10,10,8,7,10,9,7,1,1) constructed from the 9-D point by copying $x_9$ =1 to a new value $x_{10}$ =1 to make 5 pairs. The process of generation of CPC-R image for this **x** is as follows:

- Forming consecutive pairs of values from **x** = (8, 10,10,8,7,10, 9,7,1,1): (8.10), (10,8), (7,10), (9,7), (1,1).
- Filling cell (8,10) for pair ($x_1$, $x_2$) = (8,10) according to the grey scale value for the 1$^{st}$ pair (black).
- Filling cell (10,8) for pair ($x_3$, $x_4$) = (10,8) according to the grey scale value for the 2$^{nd}$ pair (very dark grey).
- Filling cell (7,10) for pair ($x_5$, $x_6$) = (7,10) according to the grey scale value for the 3$^{rd}$ pair (grey).
- Filling cell (9,7) for pair ($x_7$, $x_8$) = (9,7) according to the grey scale value for the 4$^{th}$ pair (light grey).
- Filling cell (1,1) for pair ($x_9$, $x_{10}$) = (1,1) according to the grey scale value for the 5$^{th}$ pair (very light grey).

It is a lossless visualization if values of all pairs ($x_i$, $x_{i+1}$) are not identical. The next versions of the CPC-R algorithm deal with colliding pairs and other steps in more elaborated ways. The version 2.1 for elaborated steps 3 is below.

**CPC-R version 2.1** with using **adjacent** cells in step 3:
3.1. *Set image coordinate system* (origin at upper left, or low left).
3.2. *Locate non-colliding pairs* ($x_i$, $x_{i+1}$) in the image at cell coordinate values ($x_i$, $x_{i+1}$).
3.3. *Set the starting adjacent cell* (right, left, top or bottom) for collided pairs.
3.4. *Set order of filling of adjacent cells* (clockwise or counterclockwise).
3.5. *Locate colliding pairs* ($x_i$, $x_{i+1}$) in the cell adjacent to ($x_i$, $x_{i+1}$) according to 3.3 and 3.4.

This version uses four adjacent cells (right, left, top, and bottom) and is called a ***cross filling***. Section 7 presents another option, which is called a *serpent filling*. It puts collided pairs to all 8 adjacent cells not only four cells that we use here.

**CPC-R version 2.2** with **splitting** cells in steps 3 and 4:
Step 3:
3.1. *Set image coordinate system* (origin upper left, or low left).
3.2. *Locate non-colliding pairs* ($x_i$, $x_{i+1}$) in the image at cell coordinate values ($x_i$,$x_{i+1}$).
3.3. *Split* colliding cells to vertical strips.

Step 4:

4. 1. *Assign* the grey scale intensity from black for $(x_1, x_2)$ and very light grey for $(x_{n-1}, x_n)$ for *non-colliding pairs.* Alternatively, *assign heatmap colors* for non-colliding pairs.

4.2. Assign the grey level intensity of the respective pair of values $(x_i, x_{i+1})$ to the strip for *colliding pairs*.

For example, two strips 4x8 will be assigned to a cell 8×8 for two collided pairs. Three strips 3x8, 3×8, and 2×8 are assigned for each three collided pairs. Similarly, 4-5 strips are generated for 4-5 collided pairs, respectively.

**CPC-R version 2.3** with the **darkest** intensity of colliding pairs is used in step S4:

4. 1. For all *non-colliding pairs assign* the grey scale intensity starting from black for $(x_1, x_2)$ and very light grey for $(x_{n-1}, x_n)$. Alternatively, *assign heatmap colors* for non-colliding pairs.

4.2. For *colliding pairs assign* the darkest grey level intensity of all pairs with equal values $(x_i, x_{i+1})$.

**CPC-R version 3.0** with **context** incorporated in step 5:

S5: *Put* an image of each n-D point on the top of both G mean (mean image of the training cases of class 1), and B mean (mean image of the training cases of class 2). Put both images side-by-side to form a double image.

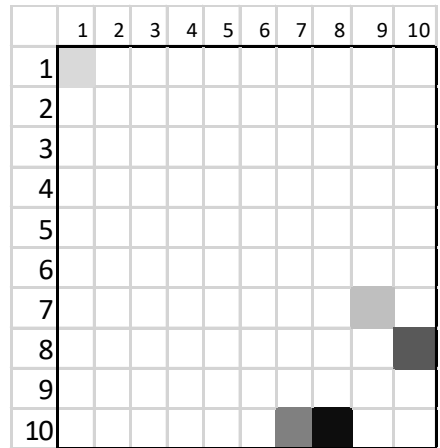

Fig.5. 10- D point (8,10,10,8,7,10, 9,7,1,1) in CPC-R.

The image size after forming a double image is (w)×(2·w). Then (2·w)×(2·w) image is formed by padding the double image. Fig. 6 shows steps of constructing a double image superimposed with mean images for each class from training data with padding. This is a way to add *context* to the images for analysis by humans and deep learning algorithms. The difference between two means in its full 34-D representation is shown in Fig. 6d *without degrading it to a lossy single distance number* to the mean n-D points.

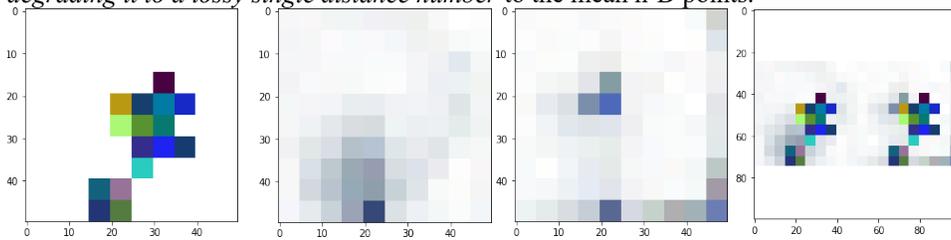

(a) Image formed after marking adjacent cells to handle collisions.

(b) G mean image – mean of all images in training set of some fold with G label.

(c) B mean image – mean of all images in training set of a given fold with B label.

(d) Final double image with image (a) on the top of mean images (b) and (c).

Fig. 6. Building double color image with context of two means for 34-D ionosphere data.

**CPC-R version 4.0** with optimized intensities and pairing of coordinates ($x_i$, $x_j$) added in step 7:

S7. *Optimize* intensities and pairing coordinates beyond sequential pairs ($x_i, x_{i+1}$) by algorithms ranged from random generating a fixed number of alternatives and testing ML prediction accuracy on training data for these alternatives to genetic algorithms. In all experiments below with data from UCI ML repository. We use 10-fold Cross Validation (CV) with such optimization.

## 2. Experiments with WBC data

Several experiments reported below have been conducted with WBC data [41] to test feasibility of the CPC-R approach(h for CNN architectures with different number of pixels per cell, which represents each pair ($x_i, x_{i+1}$). We first assign the grey scale intensities to the pixels as follows: $I(x_1,x_2)=0$ (darkest), $I(x_3,x_4)=51$, $I(x_5,x_6)=102$, $I(x_7,x_8)=153$, $I(x_9,x_{10})=204$, then we optimize these intensities and pairing coordinates ($x_i,x_j$) for accuracy in some experiments.

*2.1. Initial experiments*

Experiments **E1** and **E2** have been conducted with CPC-R **1.0** and **grey scale** values defined above. We used the CNN architecture also used in [4] that we denote as CNN64 for short and summarize below to be able to compare results with [4]:
- Convolutional layer with 64 output channels, a kernel shape of 2x2, stride of 2×2 and RELU activation.
- Convolutional layer with 64 output channels, a kernel shape of 2x2, stride of 2x2 and RELU activation.
- Pooling layer with pooling size of 2×2.
- Drop out layer with 0.4 fraction of input units to drop.
- Convolutional layer with 128 output channels, a kernel shape of 2×2, stride of 2×2 and RELU activation
- RELU Convolutional layer with 128 output channels, a kernel shape of 2x2, stride of 2×2 and relu activations.
- Pooling layer with pooling size of 2×2.
- Drop out layer with 0.4 fraction of input units to drop.
- Fully connected layer with 256 output nodes and RELU activation
- Drop out layer with 0.4 fraction of input units to drop.
- Fully connected layer with number of output nodes equal to the number of classes, with SoftMax activation.

Experiments **E2** involve the InceptionResNetV2 CNN architecture [12], which we denote as ResNetV2, with the following properties: Flatten → Dense (256, relu) → Dense(1,sigmoid). The impact of image sizes and origins of coordinates: upper left corner (ULC) or low left corner (LLC) is explored in E1 and E2 (see Table 1 with results).

Table 1. Achieved accuracies of experiments E1 and E2 in 10-fold cross validation.

| Exp. | Model | Image | Origin | Accuracy |
|---|---|---|---|---|
| E1 | CNN64 | 100×100 | ULC | 95.61 |
| E2 | ResNetV2 | 80×80 | LLC | 94.45 |

### 2.2. Experiments dealing with colliding values

The impact of colliding values along with image sizes on accuracy using ResNetV2 and LeNet5 architecture [8] is explored in experiments C1 and C2. See Table 2.

Experiments **C1** are conducted with CPC-R **2.1**, grey scale images, different orders of filling **the adjacent cells,** and two CNN architectures. *Slight increase in accuracy* was observed over E1 and E2, showing that this way to resolve collision is beneficial. We also observed that accuracy of 80x80 images is about the same as of larger 160x160 images.

Table 2. Achieved accuracies in experiments C1 and C2 with putting colliding values to adjacent cells in 10-fold CV.

| Exp. | Model | Image | Accuracy |
|---|---|---|---|
| C1 | ResNetV2 | 80×80 | 95.61 |
| C2 | ResNetV2 | 320×320 | 96.20 |
| C2 | LeNet5 | 50×50 | 94.88 |

The impact of the **splitting of colliding cells** was explored in Experiments C2 with vertical strips. C2 results are slightly better than E1 and E2, which do not address collision. The LeNet5 architecture provided 94.88% accuracy for smaller images, 50x50, than other CNNs.

### 2.3. Experiments with double images and padding

The impact of adding **context** via double images on the accuracy was explored in **Experiments D1-D8** with CPC-R **3.0** for different architectures, image sizes, and other parameters listed below for each experiment setting D1-D8. We tested the hypothesis that adding the context can increase accuracy.

While we extensively varied the parameters of experiments D1-D8, to save space Table 3 contains only results for the parameters with the best accuracies. The settings of experiments are as follows:

  D1 with superimposed colliding values (v1.0),
  D2 with filling adjacent cells (v.2.1),
  D3 with splits of cells (v 2.2),
  D4 with darkest cells for colliding pairs (v2.3),
  D5 with red levels and grey means,
  D6 with red levels and colored means,
  D7 with grey cases and colored means,
  D8 with all colored images.

The accuracy in these D experiments with double images is improved over experiments E and C with single images. The accuracy of D2 is not only slightly less than D1, but it also with much larger double images. The accuracy of D3 is less than for both D1 and D2, likely due to the smaller image size. Splitting cells necessitates larger images. The advantages of D1 and D2, which are without splitting cells, are also visible relative to D4 that has lower accuracy.

Exploring the impact of the color is the goal of the next experiments. We choose randomly intensity of each color point (3 intensities for 3 channels of a point). Thus, 15 random values for 5 cells are generated. We used images with **red** levels, **grey means** and filling **adjacent** cells in experiments **D5**. A *slight increase* of accuracy relative to the previous experiments D1-D4 was observed (see Table 3). Images with **red** levels for n-D points and **colored means** were used in experiments **D6**. Images with **grey** cases and **colored means** were used in experiments **D7**. **Color** cases and **colored means** were used in the experiments **D8** (see Fig. 6). In the experiments **D9**, we generated the images, with **red** levels, and **colored means** and **optimized intensities and pairing** of coordinates ($x_i$, $x_j$).

Table 3. Achieved accuracies of 10-fold cross validation experiments D1-D8 with double images.

| Exp. | Model  | Double image | Accuracy |
|------|--------|--------------|----------|
| D1   | LeNet5 | 60×60        | 97.22    |
| D2   | LeNet5 | 200×200      | 96.77    |
| D3   | LeNet5 | 60×60        | 95.90    |
| D3   | CNN64  | 200×200      | 95.90    |
| D4   | CNN64  | 160x160      | 96.19    |
| D5   | LeNet5 | 60×60        | 97.66    |
| D6   | LeNet5 | 60×60        | 97.80    |
| D6   | CNN64  | 100×100      | 97.36    |
| D7   | LeNet5 | 60×60        | 97.51    |
| D8   | LeNet5 | 60×60        | 97.37    |

The *optimization of pairing was most beneficial* to improve the accuracy, while the accuracy can vary due to the DL random elements as Table 3 shows. We used a simple optimization method with random generating a fixed number of perturbed alternatives and testing ML prediction accuracy on training data for these alternatives. The use of a more sophisticated methods will likely further improve accuracy due to abilities to explore more alternatives.

### 2.4. MLP experiments

The importance of using DL models vs. simpler MLP models was explored with the following Multilayer Perceptron (MLP) architecture that is simplification of CNN64:
- Layer with 64 output channels and RELU activation. This hidden layer with 64 nodes and a rectified linear activation function.
- Layer with 64 output channels and RELU activation. This hidden layer with 64 nodes and a rectified linear activation function.
- Dropout layer with fraction of input units to drop is 0.4.
- Layer with 128 output channels and RELU activation.
- Drop out layer with fraction of input units to drop is 0.4.
- Fully connected layer with number of output nodes equal to the number of classes, with SoftMax activation.

The **8 experiments** with **MLP** using single and double images allowed to achieve accuracy from 70.31% to 76.51% in 10-fold cross validation using the same setting as we used for DL models above. At first glance it clearly shows the advantages of using CNN models that gave from **94.45% to 97.8%** accuracy vs. much lower accuracy of MLP models. However, experiments with other data in Section 4 had shown that this conclusion is not so certain for those data. In some experiments MLP was a winner.

## 3. Experiments with Swiss rolls data

The goal of this experiment is to test abilities of CPC-R algorithm to classify data from two Swiss roll datasets (2-D and 3-D) [3]. These data are commonly used to test abilities of the algorithms to discover patterns of the data located on a low-dimensional manifold being isometric to the Euclidean space. Each 2-D Swiss roll data point ($x_1$, $x_2$) is represented in CPC-R images a single point (see Fig. 7a). To enrich the image, we generated an image with a "plus" centered at the points. See Fig. 7b.

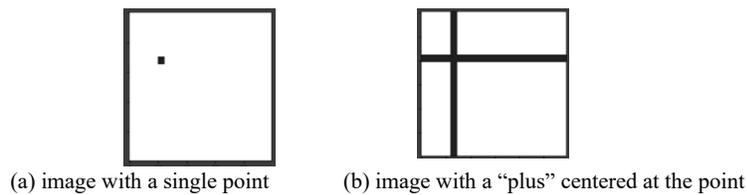

(a) image with a single point  (b) image with a "plus" centered at the point

Fig. 7. CPC-R images for 2-D Swiss roll point ($x_1$, $x_2$).

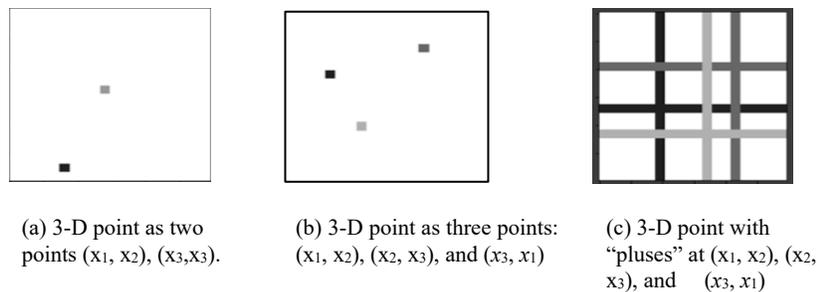

(a) 3-D point as two points ($x_1$, $x_2$), ($x_3$,$x_3$).

(b) 3-D point as three points: ($x_1$, $x_2$), ($x_2$, $x_3$), and ($x_3$, $x_1$)

(c) 3-D point with "pluses" at ($x_1$, $x_2$), ($x_2$, $x_3$), and ($x_3$, $x_1$)

Fig. 8. CPC-R images for 3-D Swiss roll point ($x_1$, $x_2$, $x_3$).

Each Swiss roll 3-D point is visualized, in CPC-R, as two 2-D points: ($x_1$, $x_2$) and ($x_3$, $x_3$). See Fig. 8a. It also can be visualized as three 2-D points: ($x_1$, $x_2$), ($x_2$, $x_3$), and ($x_3$, $x_1$). See Fig. 8b. Respectively, we generated CPC-R images, for 3-D Swiss roll, with "pluses" at three 2-D points: ($x_1$, $x_2$), ($x_2$, $x_3$), and ($x_3$, $x_1$) (Fig 4c). The max accuracies achieved are 97.87% and 97.56% for 2D and 3D Swiss rolls, respectively, in our experiments (see Table 4).

Table 4. Results for Swiss roll 2D and 3D.

| Model | Data | Image | Max accuracy by varying intensities |
|---|---|---|---|
| LeNet5 | 2-D roll | Point | 96.56 |
| LeNet5 | 2-D roll | "plus" at point | 97.88 |
| LeNet5 | 3-D roll | 2 points | 97.56 |
| LeNet5 | 3-D roll | 3 points | 94.69 |
| LeNet5 | 3-D roll | 3 "pluses" at point | 96.31 |

## 4. Experiments with Ionosphere, Glass and Car data

The setting of these experiments is the same as for WBC data experiments above with results summarized in Tables 5 and 6 for 10-fold cross validation for accuracy on the validation data. These tables report **achieved** accuracies for Ionosphere [25], Glass [37] and Car data [38], which are the best results obtained in a series of experiment of a given type. For instance, the experiments of type E2 were run for ResNetV2 for 60 and 100 epochs, images of sizes 30×30, 50×50, and 100×100, different 10- fold cross validation settings for CNN64, ResNetV2 and MLP, while Table 5 reports a single best result that is 89.98 and 50×50 images.

Table 5. Achieved accuracies of 10-fold cross validation experiments with single and double images for ionosphere and glass data.

| Ionosphere data | | | | Glass data | | | |
|---|---|---|---|---|---|---|---|
| Exp. | Model | Image size | Accuracy | Exp. | Model | Image size | Accuracy |
| E1 | ResNetV2 | 100×100 | 86.98 | E1 | CNN64 | 100×100 | 96.45 |
| E1 | MLP | 50×50 | 74.58 | E1 | MLP | 100×100 | 89.8 |
| E2 | ResNetV2 | 50×50 | 89.98 | E2 | ResNetV2 | 100×100 | **96.86** |
| E3 | MLP | 100×100 | **95.89** | E2 | MLP | 50×50 | 94.6 |
| E3 | CNN64 | 50×50 | 94.13 | E4 | MLP | 50×50 | 94.7 |
| E4 | MLP | 100×100 | 90.91 | E4 | CNN64 | 50×50 | 95.9 |
| E5 | ResNetV2 | 50×50 | 91.80 | E5 | ResNetV2 | 50×50 | 94.89 |
| E5 | MLP | 100×100 | 85.75 | E5 | MLP | 100×100 | 92.54 |
| C1 | CNN64 | 100×100 | 89.04 | C1 | ResNetV2 | 100×100 | 96.01 |
| C1 | MLP | 50×50 | 87.43 | C1 | MLP | 50×50 | 94.6 |
| D5 | ResNetV2 | 100×100 | 92.03 | D5 | CNN64 | 50×50 | 94.13 |
| D6 | CNN64 | 100×100 | 93.46 | D6 | ResNetV2 | 100×100 | 95.06 |
| D7 | CNN64 | 50×50 | 91.73 | D7 | CNN64 | 100×100 | 95.9 |
| D7 | MLP | 100×100 | 90.91 | D7 | MLP | 100×100 | 94.7 |
| D8 | CNN64 | 100×100 | 92.46 | D8 | ResNetV2 | 100×100 | 96.8 |
| D8 | MLP | 50×50 | 90.29 | D8 | MLP | 200×200 | 92.78 |

In Table 5, the best accuracy is 95.89% for the MLP classifier for Ionosphere data in E3 and the best accuracy is 96.86% for ResNetV2 classifier for Glass data in E2. This table shows that MLP was a winner only in E3 experiment for Ionosphere data being competitive in other experiments.

The car data set [38] includes 1728 instances of 4 classes with 6 attributes. The data have been normalized in [0,10] interval. We conducted multiple experiments to explore the impact of combinations of properties of CPC-R images on accuracy of classification. The Experiments E1-E7 with the Car data used MLP and CNN classifiers. Table 6 summarizes the best accuracy results with the placement of the collided pairs using *cross* filling process defined in section 1.4. These experiments involve optimization of coordinate pairing and intensities of pairs. The achieved accuracies are between 79.68% and 96.8% for the Car data. The experiments with the MLP classifier provided lower accuracies than other classifications.

Table 6. Results of E1-E7 experiments for Car data with cross filling of collided pairs in 0-fold cross validation.

| Exp. | Image | Epochs | Model | Accuracy | Exp. | Image | Epochs | Model | Accuracy |
|---|---|---|---|---|---|---|---|---|---|
| E1 | 50×50 | 100 | CNN64 | 94.79 | E4 | 100×100 | 100 | MLP | 88.97 |
| E1 | 100×100 | 100 | MLP | 83.68 | E5 | 100×100 | 150 | CNN64 | **96.8** |
| E2 | 100×100 | 100 | LeNet5 | 91.01 | E5 | 50×50 | 50 | LeNet5 | 92.03 |
| E2 | 50×50 | 100 | CNN64 | 90.60 | E5 | 100×100 | 100 | MLP | 83.87 |
| E2 | 100×100 | 100 | MLP | 80.01 | E6 | 50×50 | 50 | LeNet5 | 92.04 |
| E3 | 50×50 | 150 | LeNet5 | 96.75 | E6 | 50×50 | 50 | CNN64 | 94.08 |
| E3 | 100×100 | 100 | CNN64 | 93.99 | E6 | 100×100 | 100 | MLP | 85.87 |
| E3 | 100×100 | 100 | MLP | 87.09 | E7 | 50×50 | 100 | CNN64 | 94.07 |
| E4 | 100×100 | 100 | CNN64 | 96.28 | E7 | 50×50 | 50 | LeNet5 | 96.28 |
| E4 | 50×50 | 100 | LeNet5 | 94.63 | E7 | 100×100 | 100 | MLP | 87.91 |
| E4 | 100×100 | 100 | MLP | 88.97 | | | | | |

## 5. Saliency Maps

The goal of this experiment is exploring abilities to discover most informative attributes in CPC- R images by using the saliency maps. Specifically, we explore the approach based on the gradient $\frac{\delta output}{\delta input}$ of the output category with respect to an input image that allows observing how the output value changes with a minor change in the input image pixels. This approach expects that visualizing these gradients with the same shape as the image will produce some intuition of attention [36]. These gradients highlight the input regions that can cause a major change in the output and highlight the salient image regions that contribute towards the output [19].

*5.1 Image-specific class saliency visualization*

The mechanism of this saliency approach is as follows [20]. Pixels are ranked with a given image $I_0$, a class c, and a classification ConvNet with the class score function $S_c(I)$, influencing the score $S_c(I_0)$, e. g., a linear score model for the class c

$$S_c(I) \approx w_c^T I + b_c$$

where *I* is an image in a one-dimensional form (vectorized), $w_c$ is the weight vector, and $b_c$ is the bias of the model. Within this model each w values "defines the importance of the corresponding pixels of an image *I* for the class *c*" [20]. Then this statement is softened in [20] to accommodate the fact that in deep convolution networks, the class score $S_c(I)$ is a non-linear function of *I*. Thus, now $S_c(I)$ is considered as a linear function only locally that is computed as the first-order Taylor expansion, where w is the derivative of $S_c$ with respect to the image *I* at the point (image) $I_0$:

$$w = \frac{\delta S_c}{\delta I}$$

Next, the saliency maps are visualized with the highest-class score (top class prediction) on a test image from the data set. We conducted such saliency experiments with CNN on the randomly selected ionosphere CPC-R images to check the importance of the corresponding pixels in these images.

### 5.2. *Saliency visualization with Ionosphere data set (experiment E8)*

In this experiment computing saliency maps includes two steps:

1. Generating an image that *maximizes the class's score*, which visualizes the CNN [20,21].
2. Computing the class saliency map *specific to a given image* and class.

Both steps are based on computing the gradient of a class score to the input image. The experiment has been conducted using Keras-vis with its components Visualize_saliency and Visual_cam implemented with backpropagation modifiers [18].

Several deep learning architectures have been trained on Ionosphere data [25] to produce saliency maps. This dataset includes 351 instances of 2 classes with 34 attributes each used in 10-fold cross validation.

The list of those networks and accuracies obtained in this process of producing saliency maps are presented in Table 7. The analysis of resulting saliency maps shows that maps for all of them are extremely similar for each image as Fig. 9 illustrates.

Fig. 9 shows the representative saliency maps result for CNN64 and MLP architectures defined in Section 2 for one of the CPC-R images from the Ionosphere data: (a) original image (b) Guided backpropagation, (c) Grad-CAM, which localizes class discriminative regions, and (d) Guided Grad-CAM that combines (b) and (c). The gray scale intensities are normalized from 0 to 1. We can easily locate the dark pixels in Fig.9c (Grad-CAM).

In the Gradient Weighted Class Activation Mapping (Grad-CAM) [22, 23] the class-specific gradient information is flowing into the final convolutional layer of CNN to generate the localization map of important regions in the image. We also explored the combination of the Grad-CAM with Guided backpropagation to create a high-resolution class discriminative visualization known as Guided Grad-CAM [22].

Grad-CAM localizes the relevant image regions, but does not show finely grained importance, like pixel-space gradient visualization methods such as Guided backpropagation do. For example, Grad-CAM can easily localize the darkest pixel in the input region but not others. To combine the best aspects of both, Guided Backpropagation and Grad-CAM visualizations are fused via point-wise multiplication [22]. Guided Backpropagation was chosen over deconvolution because backpropagation visualizations being generally noise-free [24].

Fig. 10 shows saliency results for more CPC-R images from the Ionosphere data, where Fig 10a presents CPC-R images produced using the color coding. The CPC-R images in the first and last rows are from class B and CPC-R images in rows 2-4 are from class G. It is visible that these randomly selected CPC-R images from the validation set are very different without obvious visual patterns that would allow to classify them to class B or G, but trained CNN64 was able to classify them correctly.

Table 7. The results of Saliency experiment E8 with Ionosphere data.

| Model | Image Size | Epochs | Cross validation | Accuracy |
| --- | --- | --- | --- | --- |
| CNN64 | 50×50 | 50 | 10-fold | 94.4 |
| CNN64 | 50×50 | 100 | Stratified 10-fold | 94.58 |
| CNN64 | 100×100 | 50 | 10-fold | 91.6 |
| CNN64 | 100×100 | 100 | Stratified 10-fold | 91.46 |
| ResNetV2 | 100×100 | 100 | Stratified 10-fold | 93.46 |
| ResNetV2 | 50×50 | 50 | 10-fold | 95.01 |
| MLP | 50×50 | 100 | Stratified 10-fold | 88.07 |
| MLP | 100×100 | 50 | 10-fold | 90.75 |

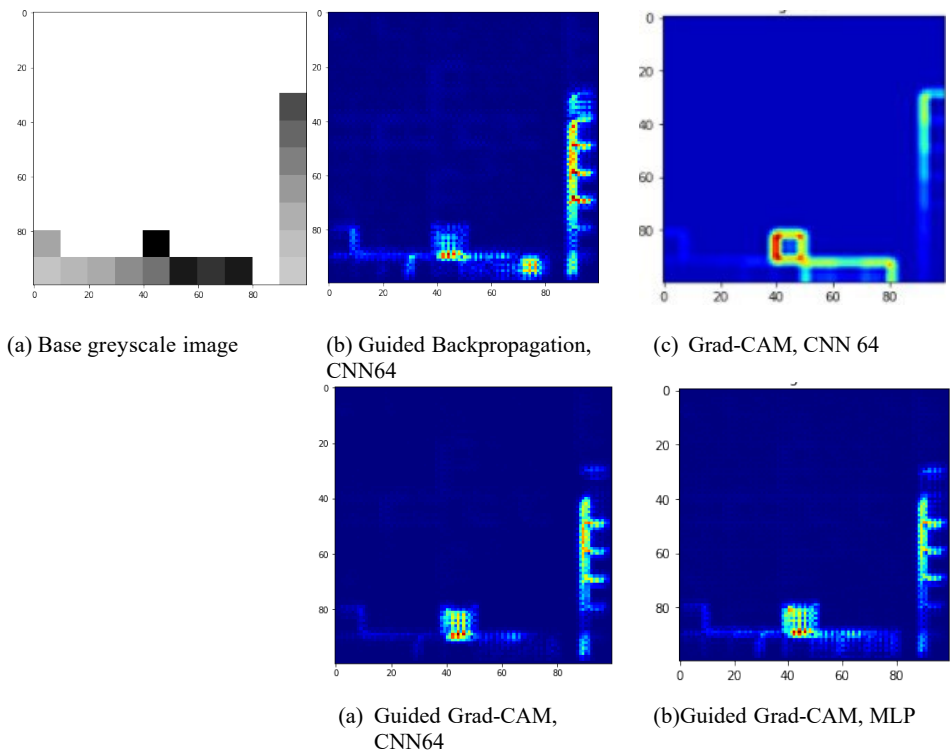

(a) Base greyscale image    (b) Guided Backpropagation, CNN64    (c) Grad-CAM, CNN 64

(a) Guided Grad-CAM, CNN64    (b) Guided Grad-CAM, MLP

Fig. 9. Saliency maps for CNN64 and MPL.

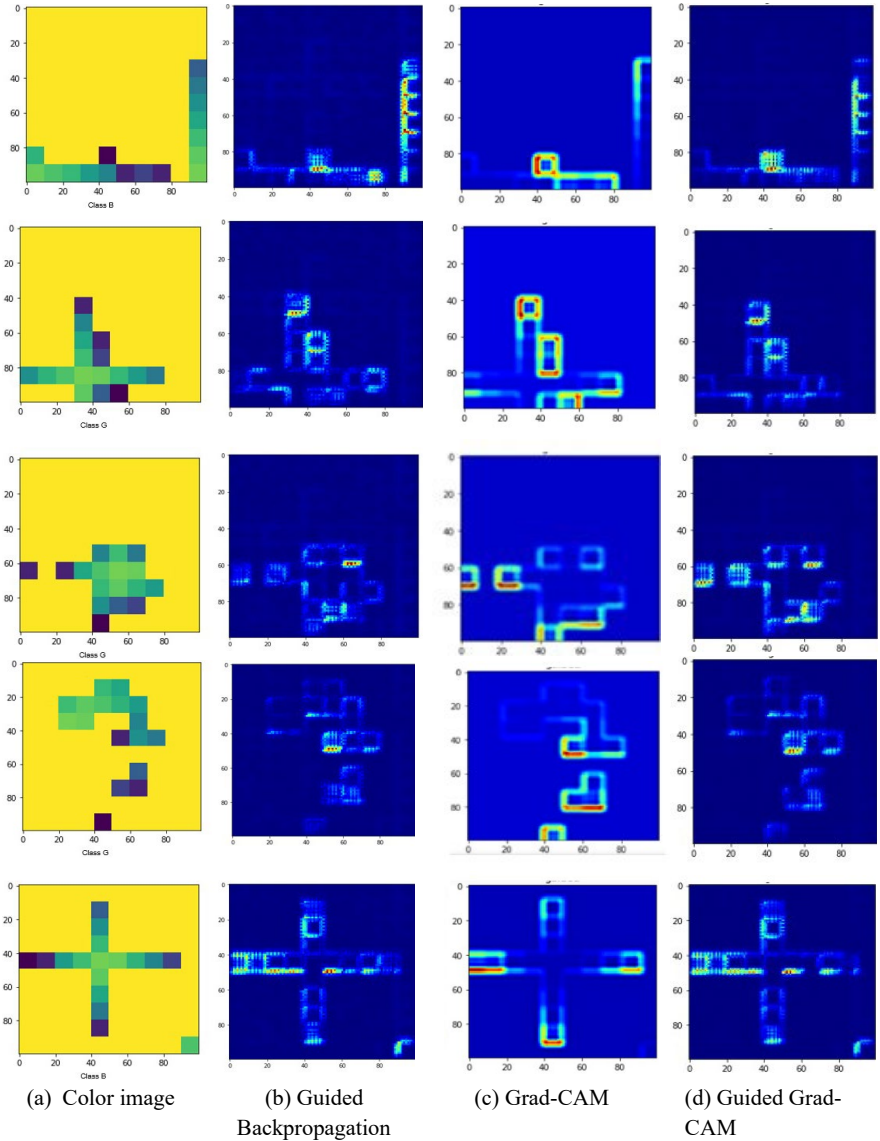

|               |                       |              |                         |
| :-----------: | :-------------------: | :----------: | :---------------------: |
| (a) Color image | (b) Guided Backpropagation | (c) Grad-CAM | (d) Guided Grad-CAM |

Figure 10. Saliency experiments with Ionosphere data.

It is visible that the most salient pixels correspond to the darkest pixels in CPC-R images. However, the saliency of these pixels is simply an **artifact** of the CPC-R coding schema where the first pair $(x_1, x_2)$ is the darkest one and the last pair $(x_{n-1}, x_n)$ is the lightest one. Thus, this saliency provides a *distorted importance* of pixels focusing only on darker start pairs like ($x_1,x_2$), ($x_3,x_4$) and others next to them. The fact that all considered CNN architectures provided quite high accuracy in 10-fold cross validation tells that those dark pairs significantly contributed to the output, but it does not justify that they are more important and relevant than others. This is illustrated in the next section where we employ an alternative method to find importance of the pixels.

## 6. Informative cells: feature importance

### 6.1. Concept of informative CPC-R cells

Experiments in the previous section shows that saliency maps did not produced the justified interpretable feature importance order on CPC-R images focusing only on dark pixels. Another common reason of saliency maps failure is their local approach with small patches without large context, and sensitivity to different contrasts [16]. Thus, we need another method that will be able to deal with these issues better. The idea of the method presented in this section is defining the importance of features by estimating the change of prediction accuracy due to the exclusion of the individual features. The large cells (super-pixels) are used as such features that can capture a larger context. This approach is in line with Super CNN approach that is a hierarchical super pixel CNN for salient object detection [16].

| | 1 | 2 | 3 | 4 | 5 | 6 | 7 | 8 | 9 | 10 |
|---|---|---|---|---|---|---|---|---|---|---|
| 1 | Cell 21 | Cell 21 | Cell 22 | Cell 22 | Cell 23 | Cell 23 | Cell 24 | Cell 24 | Cell 25 | Cell 25 |
| 2 | Cell 21 | Cell 21 | Cell 22 | Cell 22 | Cell 23 | Cell 23 | Cell 24 | Cell 24 | Cell 25 | Cell 25 |
| 3 | Cell 16 | Cell 16 | Cell 17 | Cell 17 | Cell 18 | Cell 18 | Cell 19 | Cell 19 | Cell 20 | Cell 20 |
| 4 | Cell 16 | Cell 16 | Cell 17 | Cell 17 | Cell 18 | Cell 18 | Cell 19 | Cell 19 | Cell 20 | Cell 20 |
| 5 | Cell 11 | Cell 11 | Cell 12 | Cell 12 | Cell 13 | Cell 13 | Cell 14 | Cell 14 | Cell 15 | Cell 15 |
| 6 | Cell 11 | Cell 11 | Cell 12 | Cell 12 | Cell 13 | Cell 13 | Cell 14 | Cell 14 | Cell 15 | Cell 15 |
| 7 | Cell 6 | Cell 6 | Cell 7 | Cell 7 | Cell 8 | Cell 8 | Cell 9 | Cell 9 | Cell 10 | Cell 10 |
| 8 | Cell 6 | Cell 6 | Cell 7 | Cell 7 | Cell 8 | Cell 8 | Cell 9 | Cell 9 | Cell 10 | Cell 10 |
| 9 | Cell 1 | Cell 1 | Cell 2 | Cell 2 | Cell 3 | Cell 3 | Cell 4 | Cell 4 | Cell 5 | Cell 5 |
| 10 | Cell 1 | Cell 1 | Cell 2 | Cell 2 | Cell 3 | Cell 3 | Cell 4 | Cell 4 | Cell 5 | Cell 5 |
| 0 | 1 | 2 | 3 | 4 | 5 | 6 | 7 | 8 | 9 | 10 |

Figure 11. Formation of Informative Cells.

($x_1, y_{10}$), ($x_1, y_9$), ($x_2, y_{10}$), ($x_2, y_9$) = Cell 1;
($x_3, y_{10}$), ($x_3, y_9$), ($x_4, y_{10}$), ($x_4, y_9$) = Cell 2;
($x_5, y_{10}$), ($x_5, y_9$), ($x_6, y_{10}$), ($x_6, y_9$) = Cell 3;
($x_7, y_{10}$), ($x_7, y_9$), ($x_8, y_{10}$), ($x_8, y_9$) = Cell 4;
($x_9, y_{10}$), ($x_9, y_9$), ($x_{10}, y_{10}$), ($x_{10}, y_9$) = Cell 5;
($x_1, y_8$), ($x_1, y_7$), ($x_2, y_8$), ($x_2, y_7$) = Cell 6;
($x_3, y_8$), ($x_3, y_7$), ($x_4, y_8$), ($x_4, y_7$) = Cell 7;
($x_5, y_8$), ($x_5, y_7$), ($x_6, y_8$), ($x_6, y_7$) = Cell 8;
($x_7, y_8$), ($x_7, y_7$), ($x_8, y_8$), ($x_8, y_7$) = Cell 9;
($x_9, y_8$), ($x_9, y_7$), ($x_{10}, y_8$), ($x_{10}, y_7$) = Cell 10;
($x_1, y_6$), ($x_1, y_5$), ($x_2, y_6$), ($x_2, y_5$) = Cell 11;
($x_3, y_6$), ($x_3, y_5$), ($x_4, y_6$), ($x_4, y_5$) = Cell 12;
($x_5, y_6$), ($x_5, y_5$), ($x_6, y_6$), ($x_6, y_5$) = Cell 13;

($x_7, y_6$), ($x_7, y_5$), ($x_8, y_6$), ($x_8, y_5$) = Cell14;
($x_9, y_6$), ($x_9, y_5$), ($x_{10}, y_6$), ($x_{10}, y_5$) = Cell 15;
($x_1, y_4$), ($x_1, y_3$), ($x_2, y_4$), ($x_2, y_3$) = Cell 16;
($x_3, y_4$), ($x_3, y_3$), ($x_4, y_4$), ($x_4, y_3$) = Cell 17;
($x_5, y_4$), ($x_5, y_3$), ($x_6, y_4$), ($x_6, y_3$) = Cell 18;
($x_7, y_4$), ($x_7, y_3$), ($x_8, y_4$), ($x_8, y_3$) = Cell 19;
($x_9, y_4$), ($x_9, y_3$), ($x_{10}, y_4$), ($x_{10}, y_3$) = Cell 20;
($x_1, y_2$), ($x_1, y_1$), ($x_2, y_2$), ($x_2, y_1$) = Cell 21;
($x_3, y_2$), ($x_3, y_1$), ($x_4, y_2$), ($x_4, y_1$) = Cell 22;
($x_5, y_2$), ($x_5, y_1$), ($x_6, y_2$), ($x_6, y_1$) = Cell 23;
($x_7, y_2$), ($x_7, y_1$), ($x_8, y_2$), ($x_8, y_1$) = Cell 24;
($x_9, y_2$), ($x_9, y_1$), ($x_{10}, y_2$), ($x_{10}, y_1$) = Cell 25.

The experiment below presents the accuracy of classification with covering respective cells of CPC-R images of ionosphere data that are made white. This approach views a cell as the most informative if covering it leads to largest decrease in classification accuracy. These cells are called **covered cells**. Below this method is called the **Informative Cell Covering** (**ICC**) algorithm. A 5x5 grid with 25 cells was created for each CPC-R image. Fig. 11 and the list below show this configuration. Then 25 images were created from each CPC-R image where a respective cell was made white.

## *6.2. Comparison of Guided Back Propagation salient pixels with ICC informative cells*

The covered cells, which led to the most significant drop of the accuracy, are considered as the *most informative* in the ICC approach. Table 8 shows the accuracy of classification of Ionosphere data with covered cells in the ascending order of the accuracy.

In Table 8, the lowest accuracy is 80.35% (cell 13), which is considered as the most informative in this approach. Fig. 12 shows 4 types of images for cell 13: (a) original CPC-R images from the Ionosphere data, (b) CPC-R images superimposed with heatmap when cell 13 was fully covered, (c) saliency maps for the same CPC-R image without making cell 13 white, and (d) saliency maps for the same CPC-R image with making cell 13 white.

Table 8. The classification accuracy of Ionosphere data with covered cells (images 100x100, epochs 30, 10-fold cross validation).

| Cells | Accuracy | Cells | Accuracy | Cells | Accuracy | Cells | Accuracy |
|---|---|---|---|---|---|---|---|
| 13 | 80.35 | 12 | 83.77 | 22 | 84.34 | 19 | 85.50 |
| 23 | 81.78 | 3 | 83.85 | 10 | 84.61 | 1 | 85.77 |
| 25 | 82.07 | 21 | 84.05 | 18 | 84.61 | 24 | 85.77 |
| 20 | 82.34 | 4 | 84.06 | 2 | 84.62 | 17 | 86.62 |
| 16 | 82.35 | 9 | 84.06 | 7 | 84.50 | | |
| 11 | 83.50 | 15 | 84.06 | 8 | 84.64 | | |
| 6 | 83.77 | 5 | 84.34 | 14 | 85.50 | | |

The second lowest accuracy for the ICC method is 81.78% (cell 23), which is considered as the second most informative cell. Fig. 13 contains the same four types of images as Fig. 13, but for cell 23. The highest accuracy for ICC method is 86.62% which is cell 17, which supposed to be the *least informative* cell. Fig. 14 presents four types of images for this cell.

In contrast with salient darkest cells in section 5, Figs. 12-14 (b) show that cell 13, 23 and 17 contain pairs from CPC-R images that are *not the darkest ones*, i.e., are not most salient in these images according to the Guided Back Propagation. This illustrates that ICC algorithm discovers cells that are more relevant than Guided Back Propagation in CPC-R images.

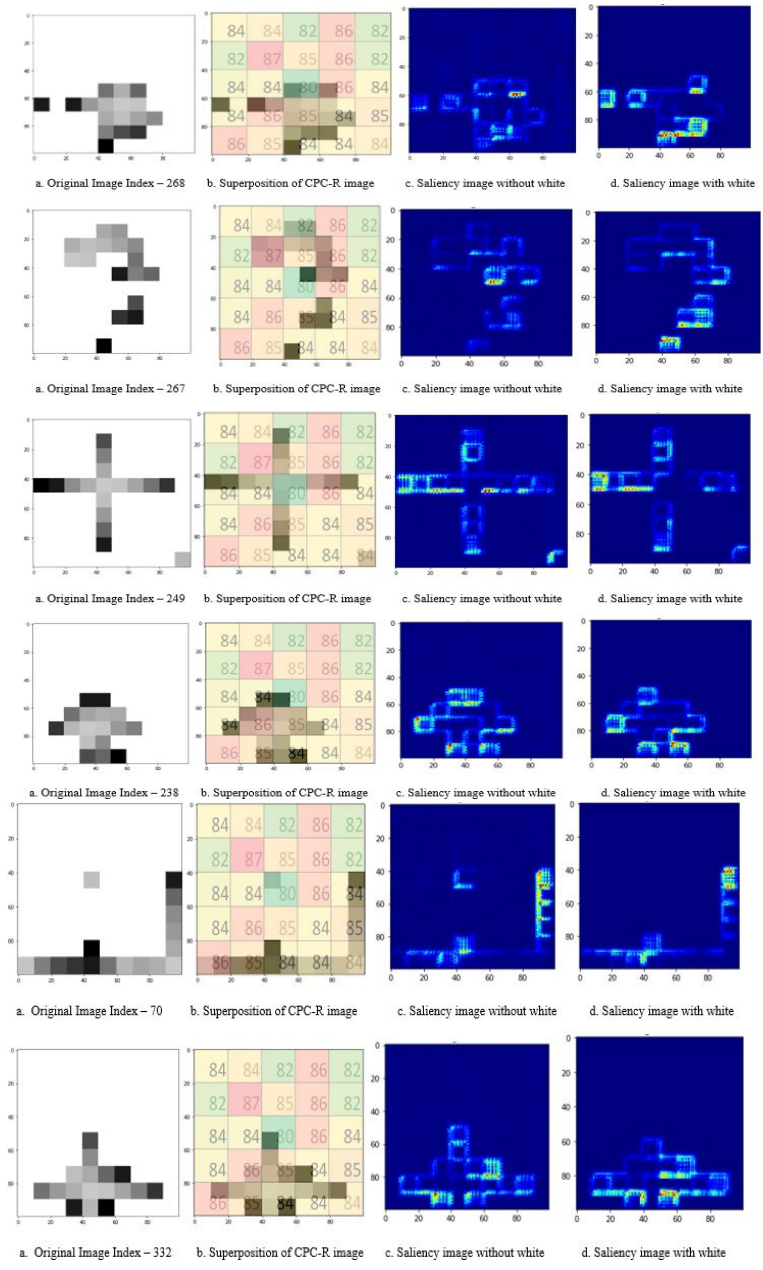

Figure 12. Informative Cell 13 (Ionosphere data).

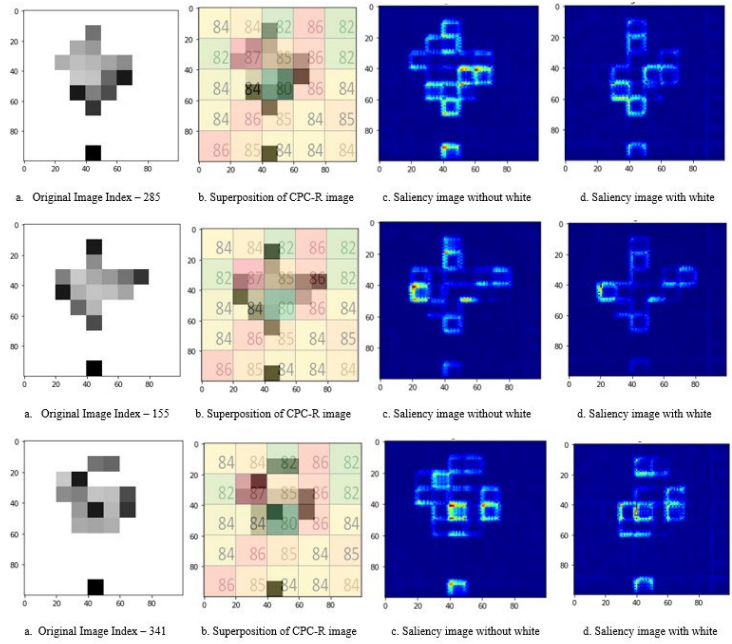
Figure 13. Informative Cell 23 (Ionosphere data).

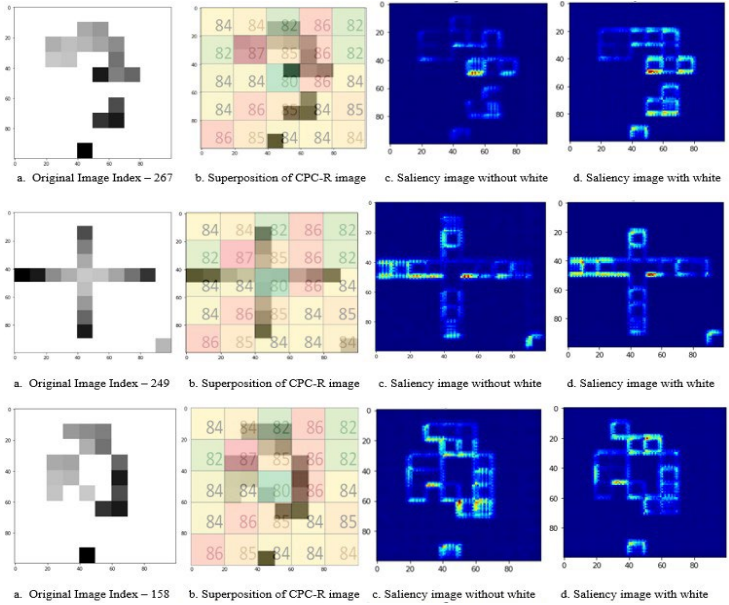
Figure 14. Informative Cell 17 (Ionosphere data).

## 6.3. Informative Cell covering with Glass and Car data

Table 9 shows the accuracy of classification of Glass [37] and Car data [38] with their CPC-R images 100x100, covered cells, 60 epochs, and 10-fold cross validation. These cells are presented in the ascending order of the accuracy. For Glass data the lowest accuracy is 86.35% for cell 18, and for Car data the lowest accuracy is 87.85% for cell 13, which are the most informative cells.

Fig. 15 illustrates informative cells for glass data and Fig. 16 summarizes informative cells for all three datasets (Ionosphere, glass, and car datasets).

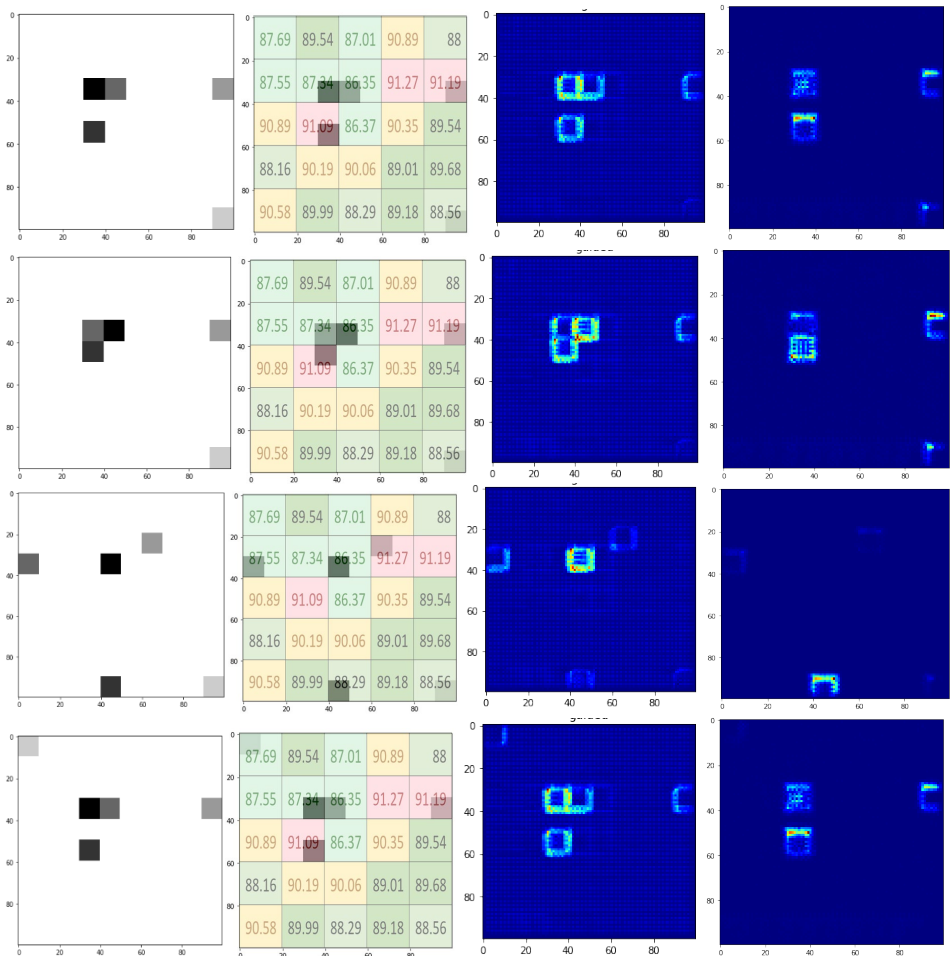

Figure 15. Glass data informative cells: a. Original image index, b. Superposition of CPC-R image, c. Saliency image without white, d. Saliency image with white.

Table 9. The classification accuracy of Glass data with covered cells.

| Glass data | | | | Car data | | | |
|---|---|---|---|---|---|---|---|
| Cells | Accuracy | Cells | Accuracy | Cells | Accuracy | Cells | Accuracy |
| 18 | 86.35 | 22 | 89.54 | 13 | 87.85 | 16 | 90.58 |
| 13 | 86.37 | 10 | 89.68 | 17 | 88.07 | 10 | 90.98 |
| 23 | 87.01 | 2 | 89.99 | 18 | 88.17 | 12 | 90.93 |
| 17 | 87.34 | 8 | 90.06 | 23 | 88.58 | 8 | 91.65 |
| 16 | 87.55 | 7 | 90.19 | 14 | 88.99 | 9 | 91.66 |
| 21 | 87.69 | 14 | 90.35 | 21 | 88.99 | 25 | 91.88 |
| 25 | 88.07 | 1 | 90.58 | 4 | 89.02 | 19 | 91.91 |
| 6 | 88.16 | 24 | 90.89 | 6 | 89.22 | 20 | 92.11 |
| 3 | 88.29 | 11 | 90.89 | 3 | 89.52 | 1 | 92.11 |
| 5 | 88.56 | 12 | 91.09 | 11 | 89.53 | 5 | 92.11 |
| 9 | 89.01 | 20 | 91.19 | 22 | 89.99 | 2 | 92.11 |
| 4 | 89.18 | 19 | 91.27 | 24 | 90.01 | 15 | 93.68 |
| 15 | 89.54 | | | 7 | 90.04 | | |

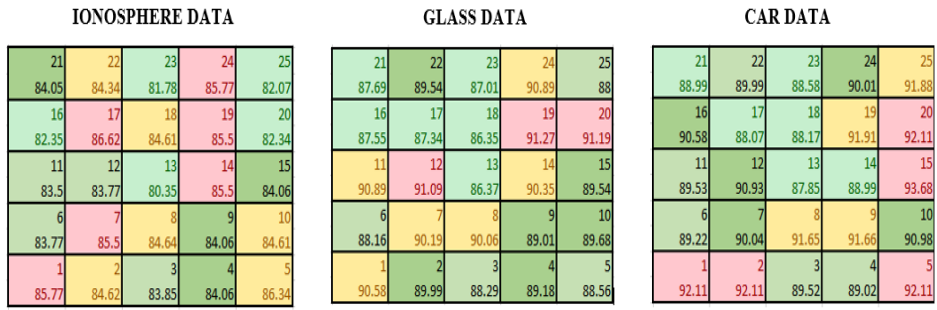

Figure 16. Informative cells.

## 7. Spiral filling of colliding pairs

This experiment with Ionosphere data uses the adjacent cells method to represent colliding pairs with the order of filling the adjacent cells that differs from the previous experiment **E2** with the same data reported in section 4. A new *spiral filling order* used in this experiment is:

right → down → left → up → lower right → lower left → upper right → upper left

Figs. 17a illustrates the method used in E2, and Fig. 17b illustrates a new spiral filling. Table 10 shows the accuracy results of the new method.

The spiral filling provided a better accuracy of 95.47% in comparison with the best results of the previous "cross" adjacent cells experiment (89.98%) in experiment E2 (see Table 5) with Ionosphere data. Table 10 presents the results of cell accuracy for each covered cell that are made white and new adjacent order. There is not much difference between the accuracies for the cells in this new adjacent order. The accuracy ranges from 88.21% to 91.15% with still cell 13 is most informative and cell 17 is among the least informative. The cells in the ascending order by accuracy values are shown in Table 11.

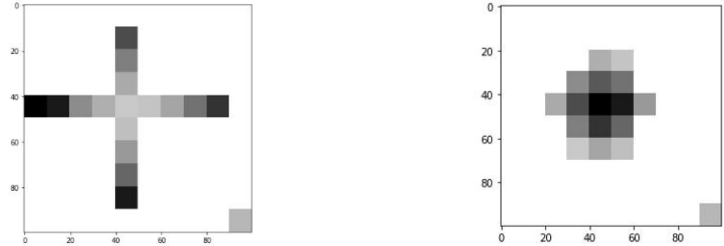

Figure 17. (a) Adjacent cells with *cross* filling, (b) Adjacent cells with new *spiral* filling.

Table 10. Ionosphere data experiment (E2 type) with spiral filling of adjacent cells.

| Image Size | Epochs | Model | Cross validation | Accuracy |
|---|---|---|---|---|
| 50×50 | 30 | ResNetV2 | 10-fold | 95.47 |
| 50×50 | 50 | CNN64 | 10-fold | 94.03 |
| 50×50 | 100 | ResNetV2 | Stratified 10-fold | 93.46 |
| 100×100 | 30 | CNN64 | 10-fold | 92.61 |
| 100×100 | 50 | MLP | 10-fold | 91.91 |

Table 11. Ionosphere data informative cells results with spiral filling of adjacent cells in 10-fold cross validation (image size 50x50, 30 epochs) for CNN64.

| Cells | Accuracy | Cells | Accuracy | Cells | Accuracy | Cells | Accuracy | Cells | Accuracy |
|---|---|---|---|---|---|---|---|---|---|
| 13 | 88.21 | 21 | 88.75 | 25 | 89.28 | 8 | 89.61 | 10 | 89.8 |
| 18 | 88.32 | 22 | 89.06 | 3 | 89.32 | 15 | 89.64 | 17 | 89.8 |
| 2 | 88.43 | 23 | 89.16 | 4 | 89.61 | 19 | 89.75 | 5 | 89.91 |
| 11 | 88.46 | 12 | 89.18 | 6 | 89.61 | 20 | 89.75 | 24 | 90.08 |
| 14 | 88.57 | 16 | 89.19 | 7 | 89.61 | 9 | 89.79 | 1 | 91.15 |

As was pointed out above the lowest accuracy for this method is 88.21% for cell 13, which is the most informative cell. Multiple experiments have been conducted to explore the impact of combinations of properties of images on accuracy of classification. The Table 12 summarizes the results. It contains only results with the different order that produced the best accuracies.

Table 12. Results of experiments of types E3-E8 for Ionosphere data with spiral filling of collided cells, 50x50 images and 50 run epochs.

| Exp. | Model | Accuracy | Exp. | Model | Accuracy | Exp. | Model | Accuracy |
|---|---|---|---|---|---|---|---|---|
| E3 | LeNet5 | 93.85 | E5 | LeNet5 | 95.85 | E8 | LeNet5 | 95.98 |
| E3 | CNN64 | 93.15 | E6 | LeNet5 | 94.54 | E8 | CNN64 | 94.56 |
| E4 | CNN64 | 93.98 | E6 | CNN64 | 93.78 | | | |
| E5 | CNN64 | **96.02** | E7 | LeNet5 | 92.98 | | | |

The above 10 results in Table 12 are best achieved accuracies produced with the spiral filling with the highest one equal to 96.02% while the lowest results in all experiments conducted in different setting is 89.98% for the ionosphere data. The experiments with the MLP classifier provided lower accuracies than other classifications (see Table 10). The above experiments involve both optimizations of the order of pairs ($x_i, x_j$) and their intensities.

## 8. Frequency for Informative Cells

In section 6 we identified the most informative cells for several datasets using the ICC method. The goal of this section is identifying which pairs of attributes ($x_i$, $x_j$) are most frequent in the most informative cells. This most frequent pairs of attributes are interpreted as **most informative pairs of attributes**.

This process of consists of the following steps:
1. Identifying pairs of values ($v_i, v_j$) of pairs of attributes ($x_i$, $x_j$) that belongs to the most informative cell.
2. Computing frequency of each pair of values ($v_i, v_j$) for each pair of attributes ($x_i$, $x_j$) that are in that cell.
3. Order frequencies in descending order for each pair of attributes ($x_i$, $x_j$)
4. Find most frequent pairs ($x_i$, $x_j$)
5. Identify pair of values ($v_i, v_j$) for the most frequent pairs ($x_i$, $x_j$).

We start showing this process for cell 13 which is the most informative for ionosphere data (see Table 11). By ICC method design each cell includes values of four pairs of attributes (see Fig. 11). For cell 13 these values are (5,5), (5,6), (6,5) and (6,6) in ionosphere data. The frequencies of these values are summarized in Table 13. It shows that most frequent and informative is pair ($x_5$, $x_6$) that appears 48 times with the most frequent pairs of values ($x_5$, $x_6$)=(6,5) and ($x_5$, $x_6$)=(5,5) that have respective frequencies 19 and 18. The next most informative pair is ($x_7$, $x_8$) that appears 40 times and the pairs ($x_1$, $x_2$) and ($x_3$, $x_4$) follow them with frequency 38.

Now we can compare the results for cell 13 with saliency results in section 5 for the same ionosphere data where the darkest pairs ($x_1$, $x_2$) and ($x_3$, $x_4$) are the most salient ones with pairs ($x_5$, $x_6$) and ($x_7$, $x_8$) follow them. Thus, in fact, the saliency map distorted the ICC importance order of attributes. It puts ($x_1$, $x_2$) and ($x_3$, $x_4$) ahead of ($x_5$, $x_6$) and ($x_7$, $x_8$) due to the fact that the CNN models were trained using "darkest first" coding schema, where ($x_1$, $x_2$) is the darkest pair. Training the models using "lightest first" coding schema, where ($x_1$, $x_2$) is the lightest pair would reverse the importance of pairs ($x_1$, $x_2$) and ($x_3$, $x_4$) putting them among least informative pairs. Thus, high salience of ($x_1$, $x_2$) and ($x_3$, $x_4$) in section 5 is rather accidental.

Note that, not only pairs ($x_5$, $x_6$) and ($x_7$, $x_8$) are most informative, but their values 5 and 6 are most informative. Thus, the Informative Cell Covering method on CPC-R images allows discovering *informative pairs* of attributes and their *specific values* not *only individual* attributes that traditional attribute covering methods do in n-D data.

Next the ICC schema and CPC-R images allow using a *less detailed data* representation than the actual measurement of the attributes as Table 14 shows. These super-pixels produced high accuracy CNN models with CNN on ionosphere data. Table 15 contains the total frequency results for the next most informative cells 14, 18 and 2.

Table 13. Ionosphere data frequency for cell 13 with and spiral order of filling adjacent cells.

| Pairs | Values | | | | | Pairs | Values | | | | |
| --- | --- | --- | --- | --- | --- | --- | --- | --- | --- | --- | --- |
| | (5, 5) | (5, 6) | (6, 5) | (6, 6) | Total | | (5, 5) | (5, 6) | (6, 5) | (6, 6) | Total |
| $x_5, x_6$ | 18 | 8 | 19 | 3 | 48 | $x_{29}, x_{30}$ | 5 | 3 | 8 | 7 | 23 |
| $x_7, x_8$ | 2 | 15 | 19 | 4 | 40 | $x_{31}, x_{32}$ | 4 | 5 | 8 | 6 | 23 |
| $x_1, x_2$ | 38 | 0 | 0 | 0 | 38 | $x_{17}, x_{18}$ | 3 | 4 | 5 | 9 | 21 |
| $x_3, x_4$ | 20 | 0 | 15 | 3 | 38 | $x_{23}, x_{24}$ | 2 | 2 | 6 | 7 | 17 |
| $x_9, x_{10}$ | 5 | 12 | 15 | 3 | 35 | $x_{21}, x_{22}$ | 1 | 2 | 4 | 7 | 14 |
| $x_{11}, x_{12}$ | 2 | 12 | 8 | 6 | 28 | $x_{15}, x_{16}$ | 4 | 0 | 6 | 2 | 12 |
| $x_{13}, x_{14}$ | 2 | 10 | 6 | 8 | 26 | $x_{19}, x_{20}$ | 0 | 3 | 2 | 6 | 11 |
| $x_{33}, x_{34}$ | 5 | 4 | 7 | 10 | 26 | $x_{27}, x_{28}$ | 2 | 1 | 3 | 4 | 10 |
| $x_{25}, x_{26}$ | 4 | 3 | 7 | 11 | 25 | | | | | | |

Table 14. CPC-R ionosphere data representation.

| CPC-R value | Actual attribute values | CPC-R value | Actual attribute values |
| --- | --- | --- | --- |
| 3 | [-0.4, -0.2) | 6 | [0.2, 0.4) |
| 4 | [-0.2, 0) | 7 | [0.4, 0.6) |
| 5 | [0, 0.2) | 8 | [0.6, 0.8) |

The *cell 14* consists of boxes with values (7,5), (5,6), (8,8) and (8,6). It's most frequent/ informative pair ($x_{15}, x_{16}$) that appears 55 times. The next most informative pair is ($x_7, x_8$) that appears 42 times. The *cell 18* consists of boxes (5,3), (5,4), (6,3), (6,4) with the most informative pair ($x_{33}, x_{34}$) that appears 43 times. The next most informative pair is ($x_{31}, x_{32}$) that appears 27 times. The *cell 2* consists of boxes (3, 9), (3,10), (4, 9), (4,10) with the most informative pair ($x_9, x_{10}$) that appears 18 times. The most informative pair after it is ($x_{11}, x_{12}$) that appears 17 times.

Table 15. Frequency Results for Cells 14, 18, and 2.

| Pairs | Cell 14 | Cell 18 | Cell 2 | Pairs | Cell 14 | Cell 18 | Cell 2 |
| --- | --- | --- | --- | --- | --- | --- | --- |
| $x_1, x_2$ | 0 | 0 | 0 | $x_{19}, x_{20}$ | 38 | 14 | 4 |
| $x_3, x_4$ | 34 | 6 | 0 | $x_{21}, x_{22}$ | 39 | 20 | 7 |
| $x_5, x_6$ | 24 | 3 | 2 | $x_{23}, x_{24}$ | 24 | 19 | 7 |
| $x_7, x_8$ | 42 | 2 | 1 | $x_{25}, x_{26}$ | 40 | 20 | 5 |
| $x_9, x_{10}$ | 39 | 8 | 18 | $x_{27}, x_{28}$ | 34 | 12 | 3 |
| $x_{11}, x_{12}$ | 34 | 10 | 17 | $x_{29}, x_{30}$ | 24 | 26 | 3 |
| $x_{13}, x_{14}$ | 29 | 11 | 7 | $x_{31}, x_{32}$ | 20 | 27 | 2 |
| $x_{15}, x_{16}$ | 55 | 7 | 7 | $x_{33}, x_{34}$ | 23 | 43 | 4 |
| $x_{17}, x_{18}$ | 27 | 15 | 8 | | | | |

In summary the results of this analysis show that most informative pairs of attributes are ($x_{15}, x_{16}$) with frequency 55 and values 7 or 8, pair ($x_5, x_6$) with frequency 48 and values 5 or 6, and pair ($x_{33}, x_{34}$) with frequency 43 times and values in 3 or 4. Thus, this algorithm shows *informative pairs* of attributes *not informative individual attributes*. It highlights *mutual dependence* of the attributes and their *joint impact* on classification accuracy.

# 9. Comparisons with other studies and CPC-R domain

## 9.1. Comparisons with other studies

The diversity of the ways to verify the results often makes the direct comparison of different methods practically impossible. Some accuracies are reported in publications without identifying its method. Other publications report 10-fold cross validation, but typically without providing actual 10 folds of the data. Different random splits of data into these 10 folds, different time limitations on model optimization lead to different accuracies. Table 18 shows comparisons of the accuracies of models presented in this chapter with other classification models.

Other variations include using the ROC curve, F measure, precision, and recall. For these reasons, we compare our results only with published 10-fold Cross Validation (CV) accuracies. Tables 16-17 summarize the comparison, while a more detailed comparison is presented below.

**WBC data**. The experiments on WBC data suggest that to get best results with CPC-R approach one needs to optimize: (1) pairing of coordinates, (2) ordering of pairs, (3) values of intensity of cells that encode pairs and (4) using the LeNet5 architecture on CPC-R images. It is a simplest among the explored network architectures that allowed the smallest size of the CPC-R images of 30x30 pixels. The obtained accuracy of 95.9-97.8% for WBC data for 10-fold CV in D1-D9 experiments suggest that D1-D9 are the best CPC-R settings to be used. There accuracies are **in the range** of the current published results: such as 94.74%, 94.36% and 95.27%, for interpretable C4.5, J4 and fuzzy decision trees, respectively, on 10–fold CV [6]. Higher accuracy from 97.97% to 99.51% on the 10–fold CV also summarized in [6] for less interpretable methods such as SVM and Neural Networks.

**Swiss roll**. Table 16 shows that CPC-R accuracies are slightly higher than obtained in [4] for 10-fold CV of CNN models on the original numeric data without transforming them to images, and on images constructed [4].

TABLE 16. COMPARISON OF ACCURACIES WITH RESULTS FROM [4].

|  | Numeric data [4] | Images [4] | CPC-R |
|---|---|---|---|
| Swiss roll 2-D | 72.50 | 97.43 | 97.87 |
| Swiss roll 3-D | 96.18 | 97.55 | 97.56 |
| WBC | 96.92 | 97.22 | 97.66 |

The reported max accuracies for 2-D Swiss roll [14] vary dramatically from 51% to 71.24% and 96.6% reported in [14] for Autoencoder, PCA and Isomap, respectively. Our accuracies are at the same level. Beyond Swiss rolls, similar spirals have been explored using DL models in [15] with reported very high accuracy reached.

Our results show CPC-R image with DL methods can *model manifolds such Swiss roll in the CPC-R visual form as images*. Thus, we expanded the methods to model and discover manifolds.

**Ionosphere data**. The range of reported accuracies for ionosphere data is from 93% to 100% on training and validation data [6] for 70% to training and 30% to validation data and 10-fold CV from different sources: 93% for MLP, 94.87% for C4.5, 94.59% for Rule

Induction, 97.33% for SVM without converting n-D data to images. We obtained 94.13% by using CPC-R that is in the range of the published results.

Therefore, the CPC-R methodology is competitive with advantages of using CPC-R images uniformly with any DL and MLP algorithm along with lossless n-D data visualization as CPC-R image.

**Glass data**. The reported results range from 68.2% for C4.5 to 98.13% for random forest [1, 5, 9] but without presenting the way of getting these numbers, or by using the area under the ROC curve and F [11] with 10-fold CV. The indirect comparison shows that our achieved 96.01% accuracy in 10-fold cross validation is also competitive and in the range of results reported in the literature.

The max accuracies for ionosphere and glass data are different. The experiments with CPC-R images resulted in accuracies between 69.8% and 95.47% for the ionosphere data and from 84% to 96.8% for glass data for different architectures. The achieved accuracy result for the saliency experiment is 94.4%.

Table 17. Comparison of Different Classification Models.

| Classification Algorithm Accuracy | Accuracy |
|---|---|
| **Breast Cancer Data** | |
| *CPC-R with Cross filling* | **95.58** |
| *CPC-R with Spiral filling* | **97.89** |
| GLC-R 5[26] | 95.61 |
| Deep Learning in Mammography [27] | 94 |
| DWS-MKL [33] | 96.9 |
| LMDT Algorithm [31] | 95.74 |
| **Ionosphere Data** | |
| *CPC-R with Cross filling* | **95.01** |
| *CPC-R with Spiral filling* | **96.02** |
| DWS-MKL [33] | 92.3 |
| ITI Algorithm [31] | 93.65 |
| Deep Extreme Learning Machine and Its Application in EEG Classification[28] | 94.74 |
| **Glass Data** | |
| *CPC-R with Cross filling* | **96.80** |
| C4.5 Algorithm [31] | 70.23 |
| Glass Classification Using Artificial Neural Network/ANN Model [30] | 96.7 |
| Comparative Analysis of classification algorithms using WEKA/MLP [32] | 67.75 |
| **Car Data** | |
| *CPC-R with Cross filling* | **96.8** |
| Performance Comparison of Data Mining Algorithms [29] | 93.22 |
| A Large-Scale Car Dataset for Fine-Grained Categorization andVerification [ 34] | 83.22 |

Experiments E1 and E2 reported above had shown better accuracy with Inception ResNetV2 architecture for both datasets. The achieved accuracy for the ionosphere data is 95% in E10 and glass data is 97% in E2, which means the CPC-R can be considered as competitive with other algorithms in accuracy. The accuracy is low with small images 30 × 30 in experiments E2 and E3, where each pair occupied a single pixel. MLP architecture provided lower accuracies than other classifications. The best of the accuracies involves both optimizations of pairs in the order of the coordinates and their intensities. In summary, the proposed CPC-R algorithm is a competing alternative to other Machine Learning algorithms as conducted experiments had shown.

### *9.2. CPC-R application domain*

In the previous section we compared accuracies obtained in this work with published in the literature showing that they are competitive. Can we expect that this competitiveness will sustain on other data? This is still the open questions for classical methods, which exist for many years. The experimental comparison on datasets is always limited in the number of experiments, size and variability of data used. Therefore, it is desirable to explore theoretical options to justify the method.

Below we explore this option based on two theorems which state that:
- Neural Networks are universal approximators of continuous functions (Universal approximation theorem), and
- Every multivariate continuous function can be represented as a superposition of continuous functions of one variable (Kolmogorov-Arnold superposition theorem [2]). By combining functions of one variable we can have continuous functions of two variables. The CPC-R pairs $(x_i,x_j)$ represent this situation. Both theorems are existence theorems not constructive ones. How can it be done constructively? One way is expanding experiments with CPC-R and CNN to new data. Another way is analyzing relevant experiments already conducted with positive results. One of them is another pairs-based method [7] known as the extended Standard Generalized Additive Models called GA2M-models. It consists of univariate terms and a small number of pairwise interaction terms. These authors announced: "the surprising result that GA2M-models have almost the same performance as the best full-complexity models on a number of real datasets".

*Is it surprising is that CNN can find patterns in CPC-R images*? CNN was associated with discovering *local features* due to the layer design that generalizes nearby pixels. CPC-R images also localize similar pairs by its design and increase the neighborhoods and CNN efficiency by (1) locating collided cells in the adjacent cells, and (2) using multiple pixels to represent each pair/cell $(x_i, x_j)$ by increasing the size of each cell of the CPC-R image. Next, often individual pairs are enough [7] and local neighborhoods not necessary. We use perturbation (see experiments D9) to ensure that important pairs are not missed. Also, weights of features assigned by CNN capture importance of features that are far away from each other. All these aspects point to the sources of CNN's success on CPC-R images.

**CPC-R domain**. *Where is the domain of application of the CPC-R methodology and current versions of the CPC -R algorithm?* It is classification tasks where the *sets of pair relations* in the data are sufficient for discovering efficient classification models by respective ML algorithms. Table 18 elaborates this statement. The base version of CPC-R 1.0 algorithm can be efficient for data where pairs $(x_i,x_j)$ for each n-D point **x** are quite unique. i.e., without equal pairs $(x_i,x_j) = (x_k,x_m)$ when $i{\neq}k$ and $j{\neq}m$, or with rare and unimportant such equal pairs. For the data where the adjacent cells for each pair $(x_i,x_j)$ in the CPC-R image are free or almost free from other pairs $(x_t,x_s)$ from n-D point **x** the version of the CPC-R algorithm 2.0 is used. For the data where the adjacent cells are occupied by other pairs the split cells version of CPC-R algorithm 2.0 is applicable. For the data where multiple pairs collide the version of CPC-R 2.0 that selects the most important one can be used with its intensity assigned to the cell.

Table 18. CPC-R use scenarios.

| CPC-R version | Use scenario |
|---|---|
| 1.0 Base version | Data without pair collision or unimportant collision |
| 2.0 Collision treatment | 2.1 adjacent cells are almost free - use adjacent cells<br>2.2 adjacent cells are occupied- split cells<br>2.3 first pair is most important- use darkest intensity |
| 3.0 Adding context | Average images of different classes are distinct |
| 4.0 Optimization | Data with non-optimal intensities and attribute pairs |

If above presented versions do not produce high accuracy, then versions of CPC-R 3.0 algorithm that adds context need to be used. It adds the background average images of different classes if they are distinct. And finally, version 4.0 is applied to optimize intensities and attribute pairs. The theoretical basis to ensure that such datasets and classification models exist provide mentioned Kolmogorov-Arnold and universal approximator theorems. The diversity of datasets used in our experiments shows that CPC-R methodology can handle variety of datasets.

## 10. Generalization of CPC-R methodology and future work

### *10.1 Generalizations*

**From CPC-R images to GLC images.** General Line Coordinates (GLC) lossless visualizations use polylines (directed graphs) in 2-D to represent each n-D point [35]. In contrast each CPC-R image encodes an n-D point using intensities of pixels or their colors without connecting nodes by edges. The same CPC-R intensity-based methodology is applicable to any GLC representation of n-D such as Shifted Paired Coordinates (SPC) [35] to produce images that we denote as **GLC-R** and **SPC-R** images, where R stands for *raster*. Respectively, this methodology contains two major steps: (1) producing GLC-R images for each n-D point and then (2) feeding these GLC-R images to CNN or other algorithms to be classified. Another aspect of generalization of CPC-R images is considering it as a part of the full 2-D Machine Learning methodology proposed in [40].

**Context generalization**. In experiments D1-D8 we used the *mean* CPC-R images of classes as a background for CPC-R images of each n-D point to add context. Let's denote this image $B_{mean}$. In this approach some background cells are occluded by the cells of the CPC-R images of the n-D point, loosing this background information. We can decrease such loss of context by computing the *differences* between CPC-R image intensities and the mean image intensities and producing the image of these differences. Let's denote this image as $B_{dif}$. It can be put side-by-side with the image $B_{mean}$ forming a new image that provide more context. Another option to add context is using *frequencies* of each pair within each class. The high frequency of some pairs of values in many images of a given class can indicate patterns specific for this class. The *frequency* of each pair within each class can be computed and the higher frequency is encoded by the higher intensity producing a version of a heatmap image. Let's denote this image as $B_{freq}$. It can be combined with $B_{mean}$ and $B_{dif}$ by putting it side-by-side with them in a joint image to be used by CNN.

**Concentrating cases of a class in local 2-D area.** When we have only two attributes (X,Y), each case of each class is a 2-D point in a standard Cartesian coordinates. The ML expectation is that points of one class will concentrate in one area and points of another class will concentrate in another area and, respectively, a linear or nonlinear discrimination function can be discovered such that the points of one class are in one subspace, say, divided by a hyperplane, from the points of another class. These points can be spread in the subspace. Respectively, the salient points can be far from each other.

Similarly, in the multidimensional situation, we expect a pattern such that all points of class 1 will be in one subspace and points of class 2 will be in another subspace. However, the Collocated Paired Coordinates (CPC) space defined in [35] may not provide this property, because the graph **x**\* of n-D point **x** can spread on the CPC space and may not reveal this closeness between n-D points in the same way as in the n-D space.

Therefore, the algorithm is needed that will concentrate cases of the same class in a local area in 2-D visualization space. Such algorithm was proposed for Shifted Paired Coordinates (SPC) in [35] where the center of the class **c** is represented losslessly as a single point in 2-D. all n-D cases that are in the n-D hypercube centered in **c**, will be in a square in SPC. This property is also true for SPC-R images.

The next question is how to assign the intensity to the center of the class, which integrates all pairs (nodes) that have different intensities. The same issue exists for other n-D points that combine two or more pairs (nodes) to a single 2-D point. For instance, consider a graph **x**\* of the n-D point **x** with three 2-D points where one 2-D point $\mathbf{p}_{1,2}$ represents two pairs $\mathbf{p}_1$ and $\mathbf{p}_2$ (two graph nodes). One option is assigning *max* or *mean* of the intensities of these nodes,

$$I(\mathbf{p}_{1,2})=\max(I(\mathbf{p}_1),I(\mathbf{p}_2)), \ I(\mathbf{p}_{1,2})= (I(\mathbf{p}_1)+I(\mathbf{p}_2))/2,$$

which is a lossy approach. Other options are: (i) *splitting* the cell assigned to $\mathbf{p}_{1,2}$ to strips of different intensities or (ii) using *adjacent cells* to put the values. Both have been done for colliding cells and the later worked better (see Sections 2 and 3).

**Saliency analysis with SPC-R images**. The CNN saliency analysis in SPC-R can be more meaningful than in CPC-R because CNN focuses on local features and SPC-R has more opportunity to localize parts of the graphs than CPC-R.

**Incomplete data with GLC-R**. Incomplete datasets are common in machine learning. One of the ways to deal with them is introducing an artificial value that can encode empty spots in the data. For instance, the attribute values can be in the range [0,10] then -1 value can be assigned to empty values in this attribute, which can be visualized.

### *10.2. Toward CPC-R specific algorithms*

*Can simpler or more specialized models substitute CNN for CPC-R images?* So far, our experiments had shown that MLP provided lower accuracy with one exclusion. In experiment E2 for Glass data and experiment E3 for Ionosphere data, the MLP provided the better accuracy than CNNs) reported in section 4 in Table 5. while further studies can find simpler MLP or other models.

The traditional deep learning CNN algorithms do well on *rich natural scenes* where convolutional layers exploit the local spatial coherence in the image. These learning algorithms search for hierarchy of informative *local groups of pixels.* In contrast, MLP ignores the spatial information by using *flattened vectors* as inputs allowing to capture non-local features. The CNN algorithms capture the *non-local relations* between features *only at the later flattened layers* of CNN.

The *CPC-R artificial images are much simpler* and with less variability than natural images. Each CPC-R image consists of a set of squares (cells) with fixed intensity of all pixels inside of each square. Thus, discovering local features in CPC-R images is simpler and the full power of deep learning algorithms seems *redundant for discovering such simpler local features*. On the other side, an algorithm for CPC-R images needs to *discover complex non-local relations* between squares in CPC-R images.

*CNN on CPC-R images* aggregates artificial intensities of pixels to produce aggregated features such as vertical and horizontal lines by using respective masks. Thus, these masks *rediscover* edges of squares that are already known. See Figures in Section 5 for the saliency maps. So, this feature discovery step can be removed or modified to make them more useful. Modified feature discovery steps can include CPC-R specialized masks. The CPC-R *specialized masks* should aggregate pairs of values $(x_i, x_j)$ and their intensities, which are artificial values in contrast with natural images, where the intensity is actual amount of light that is coming to a physical image cell.

In general, the future work is developing a specialized image recognition algorithm to discover patterns in CPC-R images. In accordance with the analysis above, the design of these algorithms should focus on *modifying* and *simplifying convolutional layers* or even removing them, but with *more complex layers that capture not-local relations*. Therefore, it was suggested to use a MLP neural network without deep learning convolutional layers at all. However, so far, it did produce a better accuracy with one exclusion as we reported above. This indicated that full removal of convolutional layers is likely should be avoided, but CPC-R specific layers need to be developed and explored. Another possible avenue of the future work is defining an optimal size of cells in CPC-R images. A preliminary hypothesis is that MLP will work better on smaller cells than CNN.

Future CPC-R specific algorithms need to aggregate the values of coordinates meaningfully by developing CPC-R relevant masks. For example, in a coordinate pair $(x_i,x_j)=(3,7)$ with intensity $I(3,7) = 0.3$, both 3 and 7 have meaning as values of $x_i$ and $x_j$, but 0.3 is just a coding that preserves ordering of the pairs of attributes. We can use another value that also preserves the order of pairs.

Commonly CNN aggregate intensities of 2×2 adjacent pixels by computing their max. For CPC-R images it is a form of generalization of artificial intensities of four adjacent pairs. Consider, an example of adjacent pairs $(x_3,x_4)=(3,7)$, $(x_5,x_6)=(4,7)$, $(x_9,x_{10})=(3,6)$ and $(x_7,x_8)=(4,6)$ with respective intensities 0.9, 0.5, 0.4, and 0.3. The max of these values is 0.9, which means that we ignore all pairs, but $(x_3, x_4) = (3.6)$. It has a meaning in natural scenes focusing on most distinct pixels, but in CPC-R images its distinction is only a result of the coding schema where the increased values of intensities are assigned to preserve the order of pairs $(x_i, x_j)$. This max pair $(x_3, x_4) = (3, 7)$ can be as informative as ignored pairs or less informative than they are. We could use the opposite order that would give priority to ($x_9$, $x_{10}$) = (3.6). Thus, for CPC-R, max aggregation does not show the most distinct pair, but a pair that is closest to the beginning of the n-D point **x**. In natural scenes, a pixel with higher intensity is the most prominent pixel in the vicinity and CNN amplifies it.

Despite of this counter-intuitive masks CNN produces a high accuracy on CPC-R images that seems like a unexplained magic. This leads us to the question of explainability of CNN and Deep Learning in general. We hope that answering this question for CPC-R in the future work will be helpful to understand the general reason of CNN and DL efficiency.

### *10.3. CPC-R anonymous ML methodology*

The CPC-R images strip detailed information of values of attributes, not showing numeric values of the attributes. To restore attribute values from CPC-R images one needs to know (1) whether the coding schema is increasing or decreasing, (2) the location of the origin of the coordinates, (3) pixel conversion schema (float and integer numbers are converted to pixel coordinates). This opened the opportunity for data anonymization for machine learning. Moreover, additional coding elements can be added specifically for data anonymization.

In the *CPC-R anonymous ML methodology*, the CPC-R algorithm converts numeric n-D data into anonymized CPC-R images and CNN builds a model on them without explicitly using numeric values of the attributes of the cases. Such images can be transferred to the SaS (Software as Service) centers for processing by powerful algorithms on the high-performance platforms.

## 11. Conclusion

This chapter had shown that the combination of CNN and MLP algorithms with CPC-R images, produced from numeric n-D data by CPC-R algorithm, is a *feasible* and beneficial ML methodology. Its advantages include *lossless visualization* of n-D data, and the abilities to add *context* to the visualization by overlaying the images of n-D points with the mean images of competing classes. The CNN was most successful in these context-enhanced CPC-R images in our experiments accompanied by optimization of pairing of attributes and pixel intensities. The CPC-R methodology allows *visual explanation* of the discovered models by tracing back to the informative pairs of attributes ($x_i,x_j$) and visualizing them using a heatmap. Data anonymization also can be accomplished by using CPC-R images that represent numeric n-D data as anonymized images, which is important for many machine learning tasks. Enhancing CPC-R methodology in multiple ways outlined in Section 10 is the goal of the future research. The Python code at GitHub can be made available upon request.